\pdfoutput=1

\documentclass[11pt]{article}

\usepackage{acl}

\usepackage{times}
\usepackage{latexsym}
\usepackage{graphicx}
\usepackage[T1]{fontenc}
\usepackage{listings} %

\usepackage[utf8]{inputenc}

\usepackage{microtype}

\usepackage{inconsolata}
\usepackage{booktabs}

\title{UICoder: Finetuning Large Language Models to Generate User Interface Code through Automated Feedback}

\author{Jason Wu$^{1,2}$ \quad Eldon Schoop$^2$ \quad Alan Leung$^2$ \\ \textbf{Titus Barik$^2$} \quad \textbf{Jeffrey P. Bigham$^2$} \quad \textbf{Jeffrey Nichols$^2$} \\
  $^1$Carnegie Mellon University \\
  $^2$Apple Inc. \\
  \texttt{jsonwu@cmu.edu} \\
  \texttt{\{eldon,alleu,tbarik,jbigham,jwnichols\}@apple.com}
  }

\begin{document}
\maketitle
\begin{abstract}
Large language models (LLMs) struggle to consistently generate UI code that compiles and produces visually relevant designs. Existing approaches to improve generation rely on expensive human feedback or distilling a proprietary model. In this paper, we explore the use of automated feedback (compilers and multi-modal models) to guide LLMs to generate high-quality UI code. Our method starts with an existing LLM and iteratively produces improved models by self-generating a large synthetic dataset using an original model, applying automated tools to aggressively filter, score, and de-duplicate the data into a refined higher quality dataset.
The original LLM is improved by finetuning on this refined dataset.
We applied our approach to several open-source LLMs and compared the resulting performance to baseline models with both automated metrics and human preferences.
Our evaluation show the resulting models outperform all other downloadable baselines and approach the performance of larger proprietary models.
\end{abstract}

\section{Introduction}

A well-known survey estimates that user interface (UI) code comprises roughly half of a user-facing GUI application’s code \cite{myers1992survey}.
However, writing UI code has traditionally been a difficult and time-consuming process that requires substantial effort and expertise.
Large language models (LLMs) present a promising solution since they are trained on large amounts of natural language text and code, which allows them to relate high-level user specifications to concrete code implementations.
Following a large-scale unsupervised ``pretraining'' phase, the model is finetuned to perform specific tasks, such as generating code, based on natural language directives in a process called instruction-tuning.
Through this process, LLMs have been trained to be proficient in conversation, develop reasoning abilities, and even use external tools~\cite{schick2023toolformer,bubeck2023sparks}. %

Nevertheless, it is still difficult for many LLMs to reliably generate syntactically-correct, well-designed code for UIs, which suggests that it is an understudied problem.
Most of the examples found in crawled web pages and repositories are not self-contained or are of low quality~\cite{gunasekar2023textbooks}. Even in curated or manually authored finetuning datasets, examples of UI code are extremely rare, in some cases making up less than one percent of the overall examples in code datasets~\cite{muennighoff2023octopack}. %
Besides being a ``low-resource'' languages \cite{chen2022transferability}, UI code and toolkits are also distinguished by their programming styles, such as frequent use of functional and reactive conventions.

In this paper, we describe an automated method for training LLMs to generate UI code from textual descriptions. We specifically focus on training models to implement UIs using SwiftUI, the official framework for popular Apple platforms, though our method would likely generalize to other languages and UI toolkits.
Instead of relying on additional external data, our approach finetunes LLMs to generate improved UI code entirely from their own previous outputs.
We first prompt an existing LLM to generate a large synthetic dataset of SwiftUI programs from a list of UI descriptions.
We then use a compiler and vision-language model~\cite{radford2021learning} to aggressively score, filter, and de-deduplicate the output samples to create a refined higher quality dataset.
By finetuning on the subset of high-scoring outputs, an improved LLM learns to generate UI code that \textit{i)} successfully compiles and \textit{ii)} is relevant to the input description.
During subsequent iterations, the improved LLM generates higher-quality datasets, which results in further performance gains.

We call our resulting model UICoder, because we originally started with the open source LLM StarCoder~\cite{li2023starcoder}. In this paper, we use the latest instruction-tuned version of StarCoder, at the time of writing, called StarChat-Beta, as our base model~\cite{Tunstall2023starchat-alpha}. We applied five iterations of our method, resulting in nearly one million generated SwiftUI programs, and trained three model variations for text-to-UI code generation.
In a series of experiments that measured automated metrics and human preference, we show that our models significantly outperform other downloadable baselines and approach the performance of much larger proprietary LLMs.

Notably, these results are particularly impressive given our models originate from StarCoder, and Swift code repositories were accidentally omitted from the training of this model~\cite{li2023starcoder}. The finetuning dataset used to create StarChat-Beta from StarCoder contains just one Swift example out of ten thousand total examples.

To summarize, the contributions of our work are as follows:
\begin{enumerate}
    \item An automated method for generating description-to-code datasets for UIs by using code compilers and vision-language models to score and filter self-generated data.
    \item  The UICoder model that generates SwiftUI implementations from natural language descriptions. 
    \item A synthetic dataset for finetuning other LLMs' UI code generation capabilities, without needing to undergo the full self-training process themselves. 
\end{enumerate}

We release the weights of our models and a synthetically generated dataset that can be used to train other LLMs for UI code generation.

\section{Related Work}

LLMs are typically trained to respond to commands (\textit{i.e.,} instruction-tuning) or perform other types of specific tasks through multiple stages \cite{ouyang2022training}: \textit{i)} a base model is pre-trained on unstructured data, \textit{ii)} the model is finetuned a dataset of human-authored responses to instructions, and \textit{iii)} the model is further ``aligned" through human feedback or ratings of output. Recently, this variations of this workflow has been successfully employed to train LLMs \cite{touvron2023llama, roziere2023code}.

For tasks that require expert knowledge such as programming, an especially time and cost-intensive part of this process is the creation of a human-labeled supervised tuning dataset and collection of human ratings.
Some efforts have led to the creation of open-source finetuning datasets via volunteer-based labeling~\cite{kopf2023openassistant,DatabricksBlog2023DollyV2}, but the datasets are comparatively limited in size and do not contain many programming-related examples. %
Other approaches have focused on creating these datasets through more automated methods, including self-curation of crawled web data~\cite{li2023self}, and mining language-code examples from code repository commit messages~\cite{muennighoff2023octopack}.
Instead of using human programmers to write thousands of programs \cite{gptlabel}, another popular approach has been to query the output of strong proprietary models to generate synthetic data, which is then used to ``distill'' a new model whose weights are known~\cite{wang2022self,luo2023wizardcoder}.
While the weights of distilled models are available for download, they have usage restrictions (e.g., non-commercial use, for research purposes only) due to conflicts with the terms-of-service of proprietary models used to train them and are limited to the performance of their ``teacher" model~\cite{kopf2023openassistant, wang2023far}. %
In our experiments, we compare models trained with all of these techniques as baselines.

Another class of approaches most closely related to ours uses automated sources of feedback for model improvement.
Reinforcement learning from AI feedback (RLHAIF) is a technique that uses another LLM to rate generated output based on a set of natural language guidelines, referred to as a ``constitution'' ~\cite{bai2022constitutional}. ReST~\cite{gulcehre2023reinforced}, an algorithm that was published during the course of our project, is most similar to our work in that it also uses a filter-then-train strategy. However, we focus specifically on scoring methods relevant to UI code generation and show that developer tools (e.g., a code compiler) and vision-language models can be successfully employed as feedback sources.

From the perspective of UI generation, there are several prior approaches related to ours.
Sketchplore~\cite{todi2016sketchplore} and Scout~\cite{swearngin2020scout} were UI prototyping/design tools that integrated a layout optimizer to generate design suggestions based on a manually-defined objective function.
Neural networks have also been used to complete partially complete layouts~\cite{li2020auto} and generate layouts ``from scratch''~\cite{li2019layoutgan} or other conditional input~\cite{li2020attribute,cheng2023play}.
In contrast to this line of work, we focus on generating compilable UI code rather than pure layouts, which results in an interactive and a more implementation-friendly format.

\section{Training Procedure}
\begin{figure}[!htb]
\centering
  \includegraphics[width=0.73\linewidth]{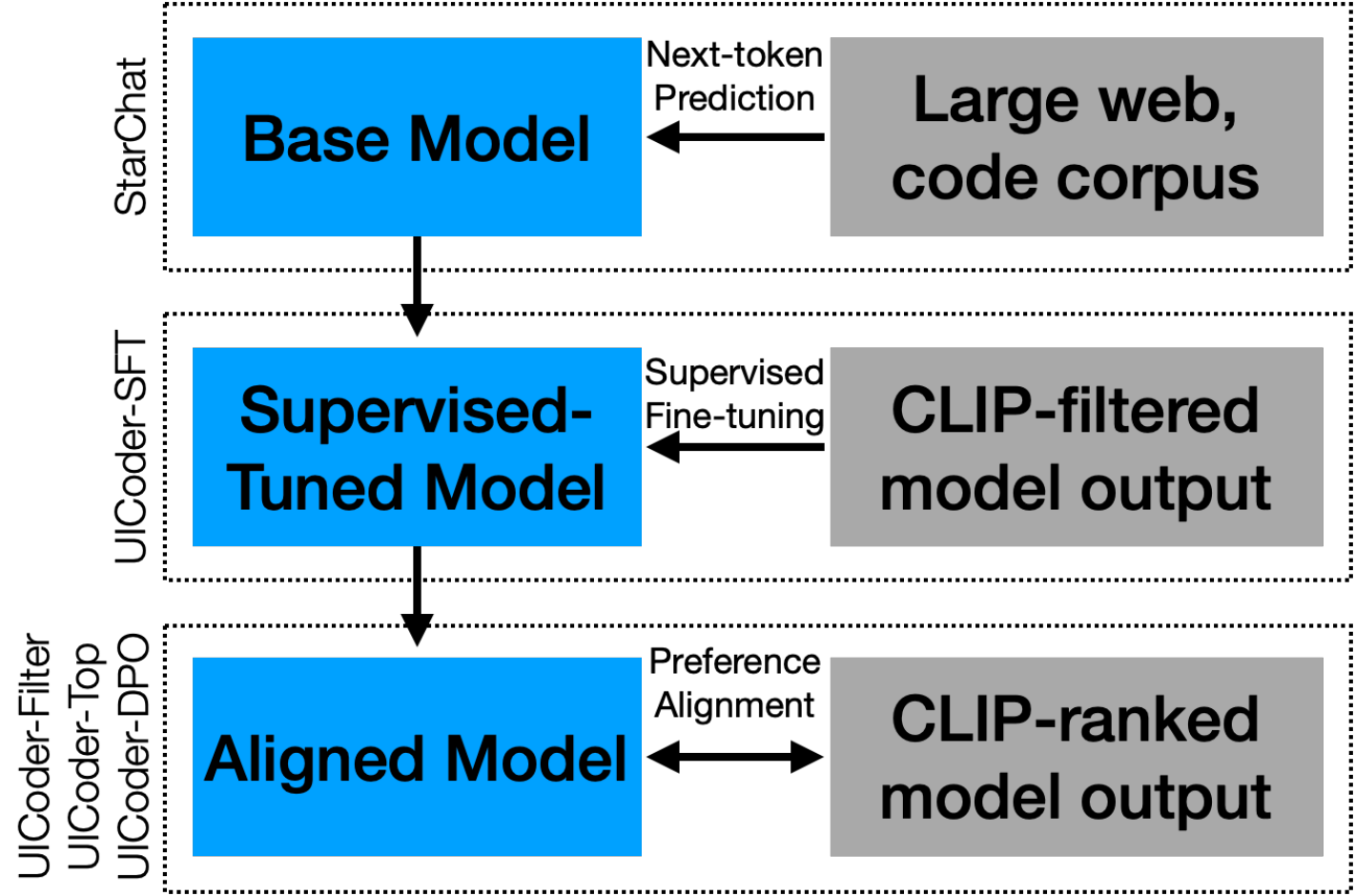}
  \caption{A flow chart showing an overview of the multi-step training process, including a base model, supervised-tuned model, and an aligned model.}
  \label{fig:overalloverview}
\end{figure}
We detail the training procedure used to train LLMs to generate SwiftUI code for a UI given its natural language description.
SwiftUI is a toolkit for the Swift language that allows cross-platform UIs (desktop, tablet, mobile, and watch) to be composed through a domain specific language (DSL).
Generating SwiftUI using LLMs is difficult, due to challenges associated with code generation in general~\cite{chen2021evaluating}, and especially poor representation of SwiftUI programs in publicly available training data and evaluation benchmarks.

To improve the generation capabilities of LLMs, we apply a training procedure based off of previous work~\cite{ouyang2022training} that involves three high-level stages.
Our approach is novel in that it uses automated feedback from code compilers and visual-language models in place of human annotations.
Figure \ref{fig:overalloverview} shows the overall overview of our training approach that involves \textit{i)} training (or using a pre-existing) base model, \textit{ii)} using supervised finetuning, then \textit{iii)} preference alignment techniques to further improve performance.
Hyperparameters were set primarily by manual experimentation, and values for various algorithms are given in the appendix.

\subsection{Training Datasets}
To train our model, we used several UI datasets: \textit{i)} Screen2Words~\cite{wang2021screen2words}, \textit{ii)} AMP~\cite{zhang2021screen}, and \textit{iii)} Crawls~\cite{feiz2022understanding}.
Between them, there are a total of ~800,000 iOS and Android UIs.
In our training procedure, we only focus on the screenshot images and natural language descriptions.
Note that these datasets do not contain the source code of UIs; therefore, the model must learn to match its own generated code output to relevant examples.
We describe two additional measures we took to increase the number and complexity of descriptions used for training.

\textbf{Generated Descriptions.}
The AMP and Crawls datasets did not contain natural language descriptions of UI screenshots, so we used a large visual-language model (VLM)~\cite{li2023blip} to weakly label these screenshots. %

\textbf{LLM-assisted Augmentation.}
The human-annotated descriptions in the Screen2Words dataset were often very simple and underspecified, and we observed that this led to simple outputs as well.
To add more detail, we used an additional open-sourced LLM~\cite{falcon40b} to paraphrase and add more detail to the original description text.
In total, we used two prompts to generate 200,000 alternate descriptions for the Screen2Words dataset.
Since the LLM used in this process did not have access to the original screenshot, it is possible that it can hallucinate inaccurate details.
We used a vision-language model \cite{radford2021learning} to compare textual descriptions and the screenshot to filter out inaccurate descriptions without a strong similarity score.

\subsection{Base Model}
We used an existing pre-trained 15B parameter model, StarChat-Beta~\cite{Tunstall2023starchat-alpha}, as a starting point, which was the best available fully-open (\textit{i.e.,} no usage restrictions) code model at the time we started the project.
StarChat-Beta is an LLM based on StarCoder~\cite{li2023starcoder} which was trained primarily on \textit{i)} TheStack~\cite{li2023starcoder}, a large dataset (250B tokens) of permissively licensed code repositories, \textit{ii)} crawled web pages, and \textit{iii)} OpenAssistant-Guanaco~\cite{dettmers2023qlora}, a small instruction-tuning dataset.
Notably, StarChat-Beta's training data contains little to no SwiftUI data~\cite{li2023starcoder}.
Swift code repositories were excluded by accident when creating TheStack dataset~\cite{li2023starcoder}, and upon manual inspection, we found that the OpenAssistant-Guanaco dataset only contains one example (out of ten thousand) with any Swift code in the response field.
We hypothesize that any Swift examples seen by StarChat-Beta during training were most likely from crawled web pages, which are possibly lower quality and less structured than repository code.

\subsection{Supervised Finetuning}

We generated input/output pairs needed for supervised finetuning by repeatedly \textit{i)} prompting an LLM to generate SwiftUI programs from a list of UI descriptions, \textit{ii)} filtering out generated programs that do not compile or have a low CLIP score, then \textit{iii)} finetuning the LLM on the high quality subset.
Our approach is similar to ``rejection sampling'' and self-training \cite{gulcehre2023reinforced} techniques, where high quality examples are mined through repeated sampling, scoring, and selection.

While the StarChat-Beta base model has poor performance overall for SwiftUI code generation, frequently producing uncompilable code or very simple UIs, we successfully detected a subset of higher-quality examples using compilation success and CLIP score as filtering conditions.
We used a randomized strategy for sampling and generating code: the model was asked to generate code for randomly sampled screen descriptions from our training datasets.
Selection was done independently, resulting in some prompts having multiple outputs and others being skipped altogether.
During each training iteration, we allocated roughly a week for sample generation, which resulted in around 100,000-200,000 total samples (influenced by average program length).

The initial proportion of mined examples is extremely small (0.4\%).
However, the model's performance improves with each iteration of training on mined examples.
As a result, it progressively generates a higher percentage of high-quality data that pass the filter.
We repeated this process four times starting with StarChat-Beta and name the resulting model \textit{UICoder-SFT}.

\subsubsection{Data Filtering}
We used three methods to filter generated data for high-quality finetuning examples: \textit{i)} compilation success, \textit{ii)} CLIP-based output relevance, and \textit{iii)} de-duplication.
Early iterations had many examples filtered out due to compilation failure since the base model was not effective at generating syntactically-correct SwiftUI code.
During later iterations, more samples were rejected due to output relevance and the de-duplication filter.
The total number of mined examples also increased over time as more generated samples fulfilled the requirements.
We also adjusted filter hyperparameters over time to increase selectivity, both to obtain higher quality training samples and to fit memory constraints.

\textbf{Compilation Success.}
We keep only fully compilable programs (compile warnings are ignored).
This is necessary because filters used in later stages rely on the rendered output of the SwiftUI code, which is not possible to generate without a working program.

\textbf{CLIP Score.}
The CLIP score filter uses a vision-language model~\cite{ilharco_gabriel_2021_5143773} to assign a numerical score to each input/output pair that measures how well the generated UI matches the input description.
We constructed a natural-language prompt template that included tags such as ``screenshot of a mobile app'' concatenated with the input description.
We used a percentile-based threshold to keep the samples with the highest scores.

Based on early observations, we noticed that CLIP occasionally produced a high score if the rendered screenshot contains the original prompt text~\cite{goh2021multimodal}.
To address this problem, we modified the CLIP score to also average the similarity score computed from the embedded ground-truth screenshot.

\textbf{De-duplication.}
We used a density-based clustering algorithm~\cite{ester1996density} to group examples based on the CLIP embeddings of their rendered screenshots.
This process makes it easier to identify groups of outputs that result in a highly similar appearance or layout which could imbalance the resulting fine-tuning set.
For each identified cluster, we keep only the example that has the highest CLIP score.

\subsection{Preference Alignment}
Following supervised finetuning, LLMs sometimes undergo an additional \textit{alignment} stage to better match with human preferences or meet specific criteria like helpfulness and harmlessness.
Unlike supervised finetuning, where the model is trained to match pre-determined outputs for each input, alignment techniques allow the model to generate its own output candidates, and then provide a numeric reward or pairwise preference labels as training signals.
We hypothesized that UICoder could benefit from this process by implicitly learning to prefer output candidates with high reward values while avoiding those with low CLIP scores or compilation errors.

\label{sec:sampleranking}
To generate data for preference modeling, we modify our data generation strategy to produce paired examples, which in our case is an input prompt paired with several possible code implementations.
After randomly selecting a screen description, we use the LLM to generate 10 outputs for the same input through a set of manually created sampling configurations.
Note that while it is possible to repeatedly sample the model with the same configuration, we found that more output diversity could be achieved with different sampling profiles.

Once multiple outputs are generated for each input, they are ranked using pairwise rules.
All compilable samples are ranked above non-compilable samples. Compilable samples are ranked by their CLIP score, and non-compilable samples are ranked by the number of error-free lines divided by the total number of lines in the program.

\subsubsection{Modeling Approaches}
We used three different modeling approaches to further finetune UICoder-SFT with the generated preference pairs, which resulted in three variations of the model: UICoder-DPO, UICoder-Top, and UICoder-Filtered.

UICoder-DPO was trained using an algorithm called Direct Preference Optimization (DPO) \cite{rafailov2023direct}.
The DPO algorithm requires significantly more GPU VRAM than supervised finetuning, so we used 4-bit quantization with the QLoRA technique so that it could fit on a single A100 GPU ~\cite{dettmers2023qlora}.
UICoder-Top was trained using a supervised training objective, but with the top output for each input as the target, which allows for greater coverage of input prompt.
Finally, as a point of comparison, we applied one more iteration of the previously used filter-then-train algorithm that ignored rankings to the newly generated data. We refer to this model as UICoder-Filter.

\section{Training Infrastructure}
We provide details about the infrastructure used to support model training. Our setup \textit{i)} generates large volumes of SwiftUI code, \textit{ii)} renders the SwiftUI code to screenshots, and \textit{iii)} scores, filters, then de-deduplicates the synthetic dataset before finetuning our model.
The servers are connected to a cloud storage provider that allows them to share files between each other.

\textbf{Code Generator.}
The code generator is used to generate a large number of SwiftUI programs using our model.
We used a mix of V100 and A100 GPUs, which was assigned by our organization's internal infrastructure.~\footnote{When using V100 GPUs, we ran the inference with the \texttt{float32} data type to support the full precision required for the dynamic range of the model's native format.}
The code generator repeatedly sampled a description of a UI from our training datasets using uniform random sampling and constructed a prompt using our model's prompt template.
A SwiftUI program was then generated by sampling our model until a pre-defined stop token was reached.
Under some training configurations, multiple outputs were generated from the same description.
Generated SwiftUI programs were periodically uploaded to network storage.

\textbf{UI Renderer.}
The UI renderer was used to convert SwiftUI programs to rendered screenshots.
The UI renderer ran macOS with Xcode and the iOS simulator that repeatedly \textit{i)} downloads queued SwiftUI programs, \textit{ii)} renders them to screenshots, then \textit{iii)} uploads the results to the network storage.
The image asset renderers ran text-to-image models that generated image assets referenced by the SwiftUI code.
We found that including generated image assets in UI screenshots was more effective than inserting placeholders (\textit{e.g.,} grey rectangles in place of images) when computing CLIP score. 

Before program compilation, we applied automated program repairs (APRs) to try to automatically fix common errors.
We focused on lightweight, manually-defined heuristic repairs (as opposed to computationally-expensive LLM-based repair methods) that used regular expressions and line numbers parsed from compiler output to match and replace errors in the source code.

\textbf{Training Server.}
The model trainer downloaded and collated the \textit{i)} input descriptions, \textit{ii)} generated SwiftUI programs, and \textit{iii)} the rendered UI screenshots.
A text-image matching model~\cite{ilharco_gabriel_2021_5143773} was used to score each rendered UI screenshot by computing the cosine similarities between \textit{i)} the embedded input description, and \textit{ii)} the embedded reference screenshot. %

Depending on the training stage, different algorithms for dataset generation and model training were applied.
All models were trained using LoRA method~\cite{hu2021lora} to improve training speed and efficiency.

\section{Experiments}
We conducted experiments to \textit{i)} measure the performance of our model over time, \textit{ii)} measure the impact of our data on different LLMs, and \textit{iii)} compare against the performance of other baselines.
\subsection{Evaluation Dataset}
We created an evaluation dataset of 200 UI descriptions, which is roughly the size of other LLM coding benchmarks~\cite{chen2021evaluating}.
We randomly selected 100 Android and 100 iOS screenshots from a held-out set.
Each screenshot was manually annotated by a member of the research team with a description that consisted of 1-3 sentences (mean 23.6 words, min 8 words, max 71 words).
Compared to screen descriptions in existing datasets~\cite{wang2021screen2words}, our descriptions were longer and contained more variation since our guidelines were less rigid (\textit{i.e.,} no constraints on sentence structure or description length).
We also used off-the-shelf UI analysis models \cite{wu2023webui} to analyze the reference screenshots in our evaluation set, which suggested they were both complex and diverse. Screens in our evaluation set contained an average of 20.3 elements per screen (min 4 elements, max 86 elements). Our evaluation set also contained many different types of UI categories and structures, \textit{e.g.,} 44 contained lists, 34 contained galleries, 23 were tutorial screens, 16 were news screens, and 14 were login screens.

For each tested model or API, we generated and rendered one program for every input in our evaluation set using the default sampling parameters and prompt template of that model, which we collected from their official web demos.
Since the purpose of the evaluation is to evaluate the generated code and layout, we replaced all image assets with the same placeholder image.
While our evaluation set doesn't contain reference code implementations, the length of model-generated output (mean length of 52 lines) also suggests higher complexity than solutions provided in other coding benchmarks \cite{chen2021evaluating} (mean 8 lines).

\subsection{Metrics}
Measuring the quality of generated UIs is challenging, and to the best of our knowledge there is no automated benchmark that can evaluate description-based UI code generation, unlike for general-purpose coding where benchmarks such as HumanEval~\cite{chen2021evaluating} may be used.
In our work, we used a combination of automated metrics and human preferences to evaluate UI code.

\textbf{Compilation Success.}
Compilation success was measured by calculating the ratio of compilable programs from the 200 input descriptions found in the evaluation set. Programs were compiled using the iOS version and Swift compiler included with Xcode 14.
A high compilation rate suggests that the model has a good mastery of the syntax needed for generating correct code.

\textbf{CLIP Score.}
Previous work has shown a correlation with human judgments when evaluating text-to-image generation~\cite{hessel2021clipscore}, and the CLIP score is more reproducible since it doesn't rely on human ratings, which may vary between people or over time.
For evaluation, we used a larger and more accurate CLIP model than the one used in our training pipeline (OpenCLIP ViT-G/14 vs ViT-B/32~\cite{ilharco_gabriel_2021_5143773}, since the memory and efficiency constraints that applied during training were not relevant during evaluation.
A high CLIP score suggests that the model's outputs are relevant to the input description.

\textbf{Human Preference Elo.}
Following recent LLM evaluation techniques~\cite{zheng2023judging}, we used pairwise human ratings to calculate Elo ratings for evaluated models~\cite{elo1967proposed}.
This method is preferred over asking annotators for absolute ratings or full rankings because it greatly reduces the cognitive load but requires more samples for a full comparison.
Each model starts out with an initial Elo rating, which is set to 1000, as done in prior work~\cite{zheng2023judging}.
For each comparison, the score of the ``winner'' was increased while the score of the ``loser'' was decreased based on the prior rating gap between the two compared models.
This method resulted in a calibrated rating score where the difference of two models' Elo ratings can be used to predict their ``winning'' probability vs. any other model.

We ran a small-scale preference-based evaluation with six human raters, who were part of the research team.
We chose to run this initial evaluation amongst the research team (HCI experts with doctorate degrees), who could apply prior knowledge/training on UI design, which would be difficult to articulate in instructions given to anonymous crowdworkers.~\footnote{We acknowledge that using a large pool of online raters could provide valuable insight from an end-user perspective.}
Note that raters were unable to see the name of the models that generated the outputs, and raters never saw any model's prior output for the evaluation prompts.
If one of the models' code did not compile, it was automatically marked as a loss, and if both did not compile, a tie was recorded.
In total, around 3000 pairwise comparisons were recorded. %

\subsection{Performance Over Time}
\begin{figure}[!htb]
\centering
  \includegraphics[width=\linewidth]{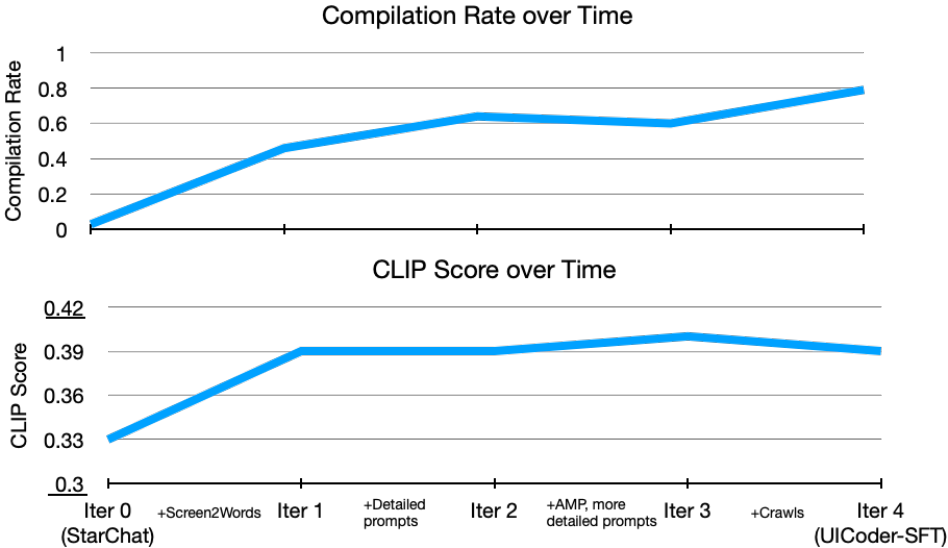}
  \caption{A plot of two automatically calculated metrics over time (on a held-out set): compilation rate and mean CLIP score. Over the course of training, our model improves metrics used to filter its training data.}
  \label{fig:performanceovertime}
\end{figure}

We measured the impact of our training procedure by running automatic evaluations of the UICoder model after each iteration, starting from the initial StarChat-Beta base model to the fourth iteration that resulted in UICoder-SFT.
The results of this experiment are shown in Figure \ref{fig:performanceovertime}.
Overall, there is a positive trend between the number of training iterations and the performance of both metrics, and the largest rate of improvement occurs during the first iteration.
The highest compilation rate was reached after the fourth iteration (0.79), while the highest CLIP Score was reached on the third iteration (0.40).

Additional UI descriptions were incorporated at various points in the training process based on our manual observations and tests.
For example, at the initial iteration from 0 to 1 we started with only human-authored descriptions found in Screen2Words~\cite{wang2021screen2words}.
However, we noticed that the model tended to produce relatively simple outputs and struggled to understand more detailed descriptions, which led us to include LLM-augmented and paraphrased descriptions.
Similarly, we later included iOS screenshots, since the Android UIs in Screen2Words~\cite{wang2021screen2words} and RICO~\cite{deka2017rico} use different design patterns and UI components.
These changes likely contributed to some of the fluctuations in model performance measured over the course of training.

We focused our evaluation on mobile app UIs, which we acknowledge only reflects a small portion of overall UIs in use and platforms supported by SwiftUI. Our choice was based on the limited selection of datasets for UIs, and we are only aware of one dataset that contains paired text descriptions \cite{wang2021screen2words}. We expect that our method could work for other platforms as well, such as web pages \cite{wu2023webui}, although the descriptions would need to be synthetically generated.

\subsection{Distillation Experiments}
We trained the UICoder models through a multi-iteration training process. However this is time-consuming because it requires repeatedly generating, evaluating, and training on self-generated data.
Therefore, we explored the possibility of using data generated during UICoder's training to finetune other LLMs, without needing to repeat the self-generation process.
Model distillation refers to the practice of using the results of a larger ``teacher model'' to finetune or train a smaller ``student model.''
This practice is employed to use the often superior results of proprietary models to train other more open models, but can also be used to transfer skills from a smaller purpose-built model to a larger general model through reverse-distillation.
Octocoder is a more recent version of our base model, StarChat-Beta~\cite{muennighoff2023octopack}. Octocoder was released after we began our model training, so we investigate the possibility of ``rebasing'' UICoder onto an improved model.
MPT-30B-Instruct~\cite{MosaicML2023Introducing} is a 30B LLM that is double the size of UICoder and was finetuned on a collection of permissively-licensed general-purpose instruction-following datasets.
MPT-7B-Instruct~\cite{MosaicML2023Introducing} is a smaller 7B version of MPT-30B-Instruct that is half the size of UICoder and might be useful for efficient deployment.

We used training examples generated from the last iteration of UICoder training to distill these models using the same hyperparameters.
We refer to the resulting models as Octocoder++, MPT-30B++, and MPT-7B++.

\textbf{Results.}
Both initial and distilled model performance were highly dependent on the size of the base model, where larger model sizes led to better performance.
The best-performing model was MPT-30B-Instruct which had a compilation rate of $0.14 \rightarrow 0.78$ and CLIP score of $0.351 \rightarrow 0.401$.
This was followed by MPT-7B, with a compilation rate of $0.13 \rightarrow 0.69$ and a CLIP score of $0.350 \rightarrow 0.395$.
Although Octocoder was a 15B model trained on code, it had the worst performance with a compilation rate of $0.06 \rightarrow 0.51$ and a CLIP score of $0.235 \rightarrow 0.382$.
This may have been due to a mismatch between Octocoder's training data, which was based on scraped commit messages, and the types of UI descriptions relevant to our task.
Overall, all models' compilation rates and CLIP scores were greatly improved by distillation, which suggests the utility of UICoder-generated data for finetuning other models.

\subsection{Baseline Comparison}
We compared the performance of UICoder models, distilled models, and several classes of baselines.

\textbf{Baseline Models.}
We categorized baselines into three categories: \textit{i)} proprietary models, \textit{ii)} restricted models, and \textit{iii)} permissive models.
We evaluated two baselines for each category.

Currently, proprietary models have the best performance, but are accessible only through web API requests and often have usage restrictions. We included GPT-4 and GPT-3.5-Turbo as proprietary baselines, since they have been shown to excel at a wide range of tasks, including code generation~\cite{bubeck2023sparks,luo2023wizardcoder}.

Restricted baselines are freely downloadable models with license or usage restrictions (\textit{i.e.,} no commercial use), due to the use of a proprietary model in generating training data.~\footnote{We do not consider some models, such as LLaMA variants, to be freely downloadable because they require a license application to first be approved by Meta to access the weights.}
We included WizardCoder~\cite{luo2023wizardcoder} and MPT-30B-Chat~\cite{MosaicML2023Introducing} as restricted baselines.
Note that MPT-30B-Chat is different from MPT-30B-Instruct, in that the chat model was finetuned using output from ChatGPT and GPT-4.

Finally, we included StarChat-Beta~\cite{Tunstall2023starchat-alpha} and OctoCoder~\cite{muennighoff2023octopack} as permissive baselines, which were both trained on permissively-licensed code repositories and instruction-tuning datasets.

\begin{table}[!htb]
\centering
\caption{Table of automated metrics and Elo ratings computed for each model on the evaluation set. We display the mean CLIP score.}
\small
\begin{tabular}{@{}lllll@{}}
\toprule
Model              & Params & Compile & CLIP         & Elo  \\ \midrule
GPT-3.5-Turbo      & -      & 0.88        & $0.416$ & $1224$ \\
GPT-4              & -      & 0.81        & $0.419$ & $1189$ \\
WizardCoder        & 15.5B  & 0.23        & $0.393$ & $870$ \\
MPT-30B-Chat       & 30B    & 0.27        & $0.368$ & $873$ \\
StarChat-Beta      & 15.5B  & 0.03        & $0.334$ & $773$ \\
Octocoder          & 15.5B  & 0.06        & $0.235$ & $777$ \\ \midrule
UICoder-Filtered   & 15.5B  & 0.79        & $0.404$ & $1099$ \\
UICoder-Top & 15.5B  & 0.82        & $0.396$ & $1084$ \\
UICoder-DPO      & 15.5B  & 0.75        & $0.393$ & $1091$ \\ \midrule
MPT-7B++  & 7B     & 0.69        & $0.395$ & $1015$ \\
Octocoder++        & 15.5B  & 0.51        & $0.382$ & $959$ \\
MPT-30B++ & 30B    & 0.78        & $0.401$ & $1047$ \\ \bottomrule
\end{tabular}
\label{tab:results}
\end{table}

\begin{figure}[!htb]
\centering
  \includegraphics[width=0.95\linewidth]{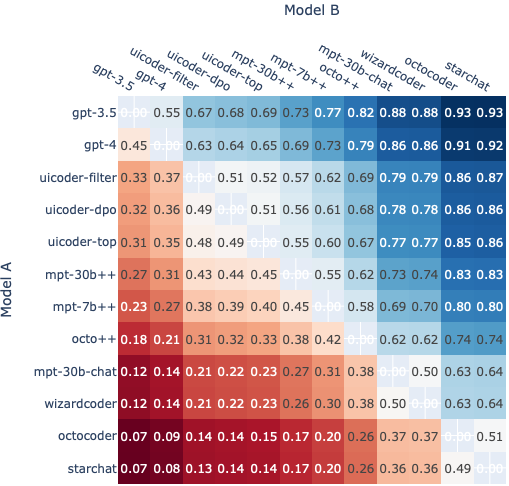}
  \caption{Matrix shows the predicted win probability of model A against model B. Our training technique significantly improved the performance of an initially poorly-performing base model (StarChat) to competitive among larger proprietary models (UICoder). }
  \label{fig:winrate}
\end{figure}

\textbf{Results.}
We ran all three metrics for the baseline comparison experiment. The results are summarized in Table \ref{tab:results} and Figure \ref{fig:winrate} shows the expected human ratings for every pair of evaluated models. 
Overall, the proprietary models had the best performance, followed by UICoder models, then models distilled from UICoder data.
It was somewhat surprising that GPT-3.5 had a higher compilation rate than GPT-4, since GPT-4 is often considered a stronger model.
Upon manual inspection, we hypothesized that GPT-4 often tried to produce longer and more complex code implementations, which made compilation errors slightly more likely.

Variations of the UICoder models approached the performance of the proprietary GPT models, and UICoder-Top had a higher compilation rate than GPT-4.
All three UICoder variants had roughly the same level of performance, which suggests that the additional preference alignment stage did not lead to significant improvements, possibly due to sub-optimal hyperparameters or the additional model quantization needed to run the training algorithm.

Distilled models finetuned from the final iteration of generated data had slightly lower performance than the UICoder models themselves, which together with our performance over time experiments (Figure \ref{fig:performanceovertime}) suggests that multiple training iterations are needed to maximize performance.
MPT-30B++ was the best-performing distilled model, possibly due to its larger model size, and although Octocoder++ is pre-trained on large amounts of code, it performed worst out of the distilled models. MPT-7B++ maintains relatively high performance with a much smaller size, which is encouraging from an efficiency standpoint.

All other downloadable models fared considerably worse.
Restrictive models performed better than permissive ones, due to the use of larger amounts of finetuning data generated from proprietary model APIs.
It is possible that the distilled proprietary data did not contain many SwiftUI-related coding examples and querying proprietary models specifically for more UI-related tasks could further boost performance.~\footnote{We did not distill SwiftUI models ourselves from proprietary LLMs because this would require breaking their ToS.}
Notably, StarCoder-Beta, our base model, was originally ranked in last place, likely due to the scarcity of Swift and SwiftUI code in its training code.
Our results suggest that our method is highly effective at improving its capabilities, since UICoder becomes one of the top performing models after training.

\section{Conclusion}
In this paper, we introduced a novel method that uses automated tools such as a code compiler and a pre-trained vision-language model to finetune LLMs to generate UI code from user-provided textual descriptions.
Unlike other instruction-following LLMs that are trained on expensive human feedback or output from a stronger proprietary model, our technique trains LLMs entirely using high-quality examples mined from self-generated data.
We applied five iterations of our algorithm to an existing open-source LLM, resulting in nearly one million generated SwiftUI programs, which were then filtered and refined to train the UICoder model.
In a series of experiments that measured both automated metrics and human preferences, UICoder and other models trained with our method outperformed all other downloadable baselines by a large margin and approached the performance of larger proprietary models.

\section{Limitations and Risks}

We discuss limitations of our training approach, specifically drawbacks of our automated feedback sources, and model evaluation. 

\textbf{Focus on SwiftUI.}
We chose to work with the SwiftUI toolchain in our work. Other toolchains with similar properties exist (\textit{e.g.,} Dart/Flutter for Google/Android and React Native), and we expect that our method could be directly applicable as well.
Our use of a compiler as automated feedback into the training process can be more broadly useful in learning the unique syntax of specialized languages and libraries.

\textbf{Reliance on Synthetic Datasets.}
UICoder’s reliance on synthetic, self-generated fine-tuning avoids the cost of human labeling/ranking and avoids violating the ToS of stronger proprietary models used for distillation. However, we acknowledge that a tradeoff of the self-generation+filtering strategy is limited diversity of fine-tuning data (since no additional external data is provided). It should be noted that our evaluation uses an unbiased, randomly sampled set of UI descriptions, and it shows that our method performs favorably against other models trained on external sources of data, which suggests effectiveness for real-world use-cases.

\textbf{Swift Compiler.}
The Swift compiler provides ground truth for compilation success, but this only returns a sole binary value, which equates the quality of any two non-compiling programs regardless of the number of defects.
Based on our observations, this can incentivise the model to avoid errors, rather than learn how to correct them, resulting in very simple outputs.
We attempted to rank generated programs by the proportion of error-free lines of code to total program length but found that this actually led to decreased a compilation rate in the case of the UICoder-DPO model.
A more useful signal would be to count the number of changes needed to ``fix'' a failing program, but this cannot be directly computed from compiler output.
As a result of these limitations, UICoder still generates relatively simple programs, and we plan to employ more advanced PL or verification techniques to improve this aspect of our approach in the future.

\textbf{CLIP Score.}
We found that CLIP could detect overall output relevance and large disparities in output quality, but it wasn't well-equipped for comparing subtle design choices or detecting some types of visual defects.
Figure \ref{fig:failurecases} shows our trained model still exhibits errors that are uncharacteristic of human developers, including data formatting, text-overflow, inaccessible controls, poor style, and contrast.
We hypothesize that a major reason for CLIP's limitations is the small fraction of its training examples that included UI screenshots.
The CLIP model is also known to struggle with certain tasks such as object-counting and reasoning~\cite{radford2021learning}, which could be relevant for some types of prompts, e.g., ``a login page with three buttons stacked vertically.''
Since CLIP only accepts image input, it can only evaluate a dynamic, interactive UI with a screenshot of its default state.

\textbf{Evaluation.}
Our evaluation is limited in that we asked human evaluators to judge UIs purely through screenshots, due to the complexities of installing the compiled app on their device. This may bias the evaluation towards aesthetic qualities of the UI rather than its functionality of navigation.
As stated, our human evaluation consisted of a limited number of expert raters, who were also a part of the research team. While measures were taken to prevent bias in rating results, we expect that a larger pool of anonymous raters could give insight from the end-user's perspective. We plan to build the infrastructure required for larger-scale tests.
We also included proprietary model endpoints such as OpenAI GPT-3.5 and GPT-4 in our baseline evaluation, which was conducted August-September 2023.
Because OpenAI periodically updates these models, this could affect reproducability.
We also didn't directly evaluate code quality, which is important for practice use.
We provide sample generations in supplementary material and leave more rigorous evaluation to future work.

\textbf{Risks.}
UICoder is an LLM model that is, in theory, capable of generating potentially problematic or harmful outputs if prompted to do so.
Our focus is to improve an existing LLM's ability to write better UI code, and should be combined with additional techniques (\textit{e.g.,} alignment and human-in-the-loop correction) to reduce this risk. Since UICoder is a derivative of a pre-trained code LLM trained on online code repositories, it could also generate code with security vulnerabilities.
Additional care should be taken to check generated code before including it. Part of our training pipeline involves automatically executing LLM-generated code, but security risks can be mitigated by running code in a virtualized, isolated environment.

\bibliography{anthology,custom}

\appendix

\section{Output Samples}
\begin{figure*}[!htb]
\centering
  \includegraphics[width=\linewidth]{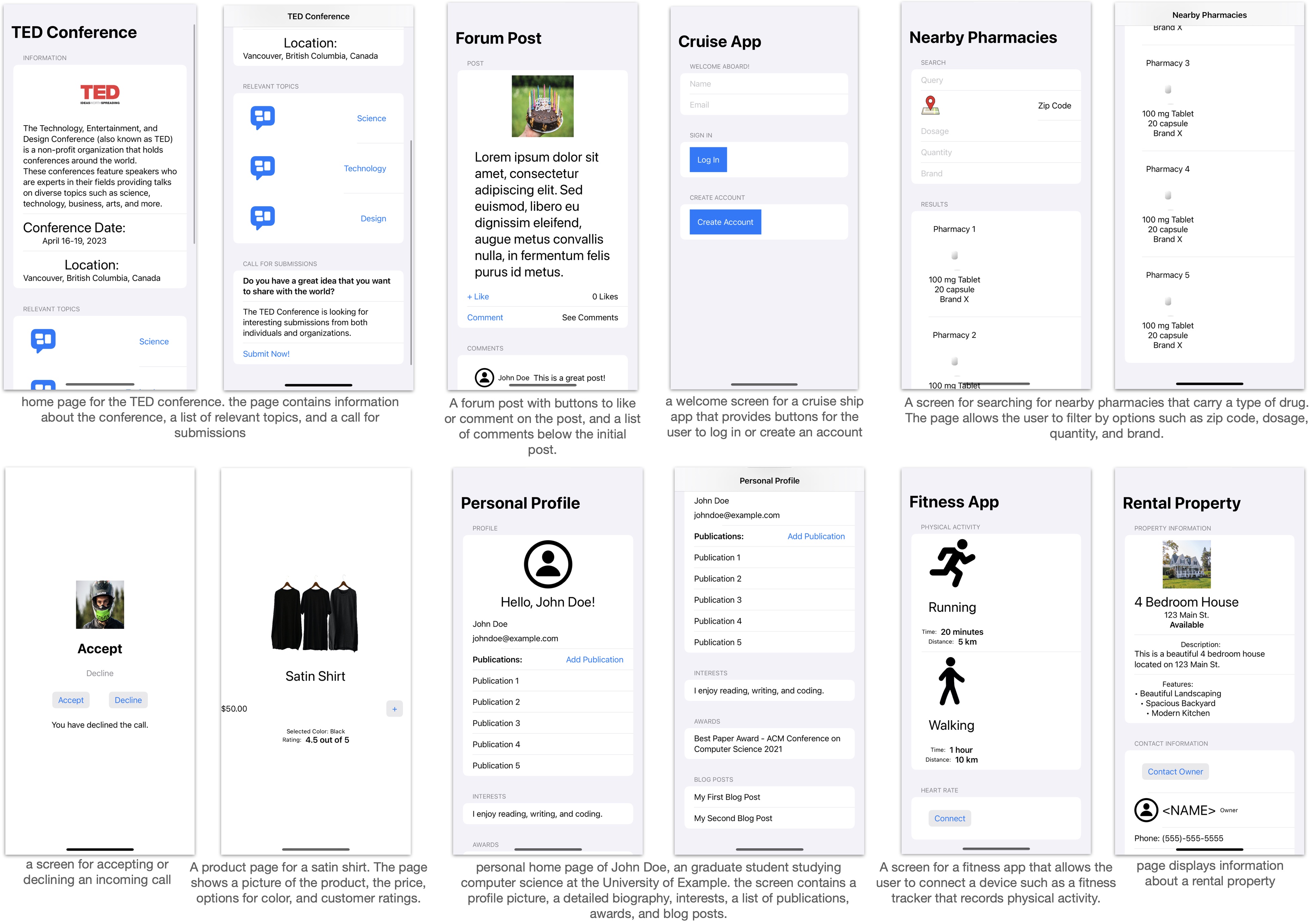}
  \caption{Screenshots rendered from SwiftUI code generated by our models. For illustration purposes we manually included stock photos and icons. The model-generated code was not modified in any way except to update image asset names.}
  \label{fig:outputsamples}
\end{figure*}

\begin{figure*}[!htb]
\centering
  \includegraphics[width=0.8\linewidth]{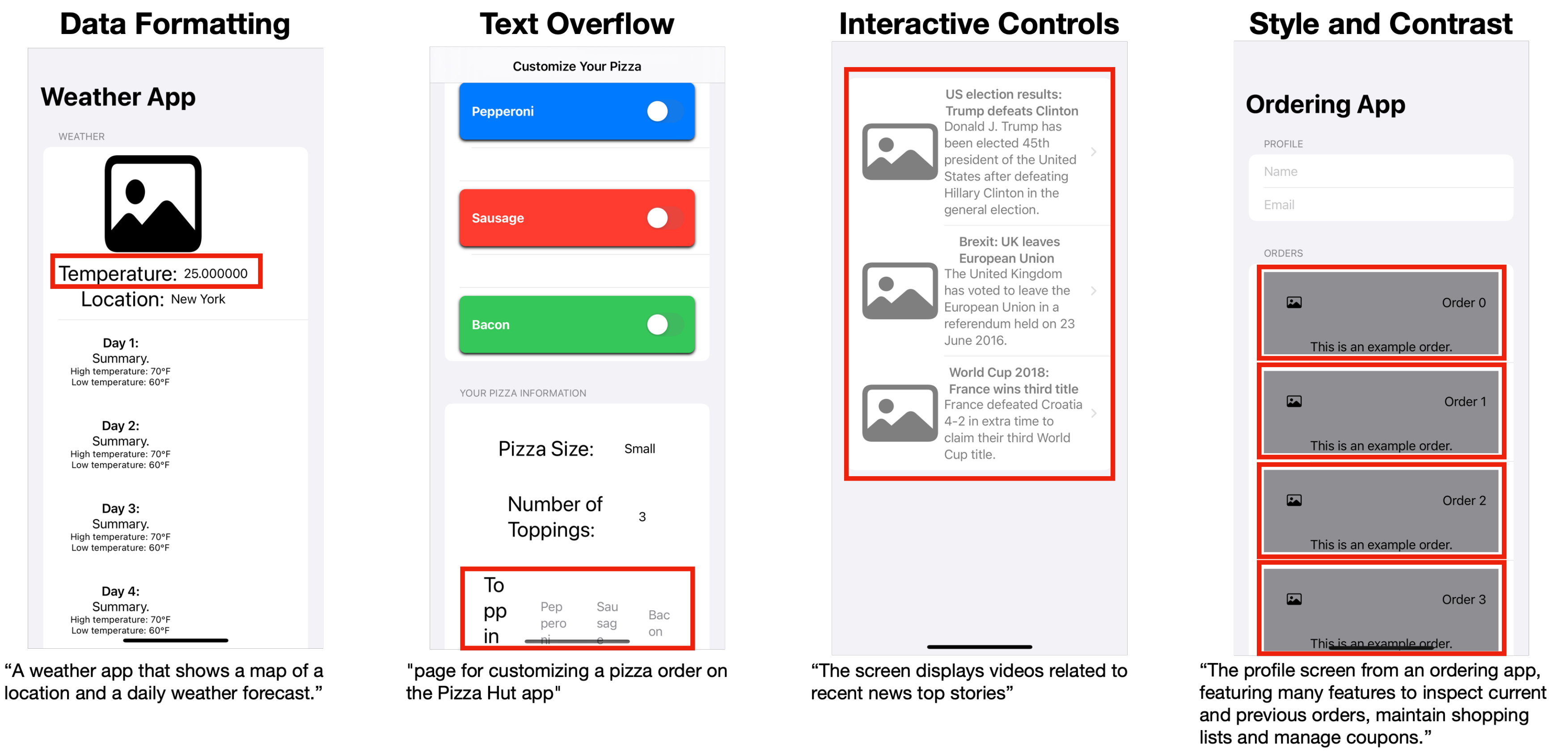}
  \caption{We demonstrate limitations of our approach through four types of failure cases observed in generated data. Note that all icons and images in these samples were replaced with placeholders.}
  \label{fig:failurecases}
\end{figure*}

Figure \ref{fig:outputsamples} shows several input descriptions and the corresponding rendered UI. Figure \ref{fig:failurecases} shows several examples containing failure cases. In addition, we include some examples of generated code in the supplementary material.
\label{sec:appendix_output}
\section{Human Evaluation Instructions}
\begin{figure*}[!htb]
  \includegraphics[width=0.9\textwidth]{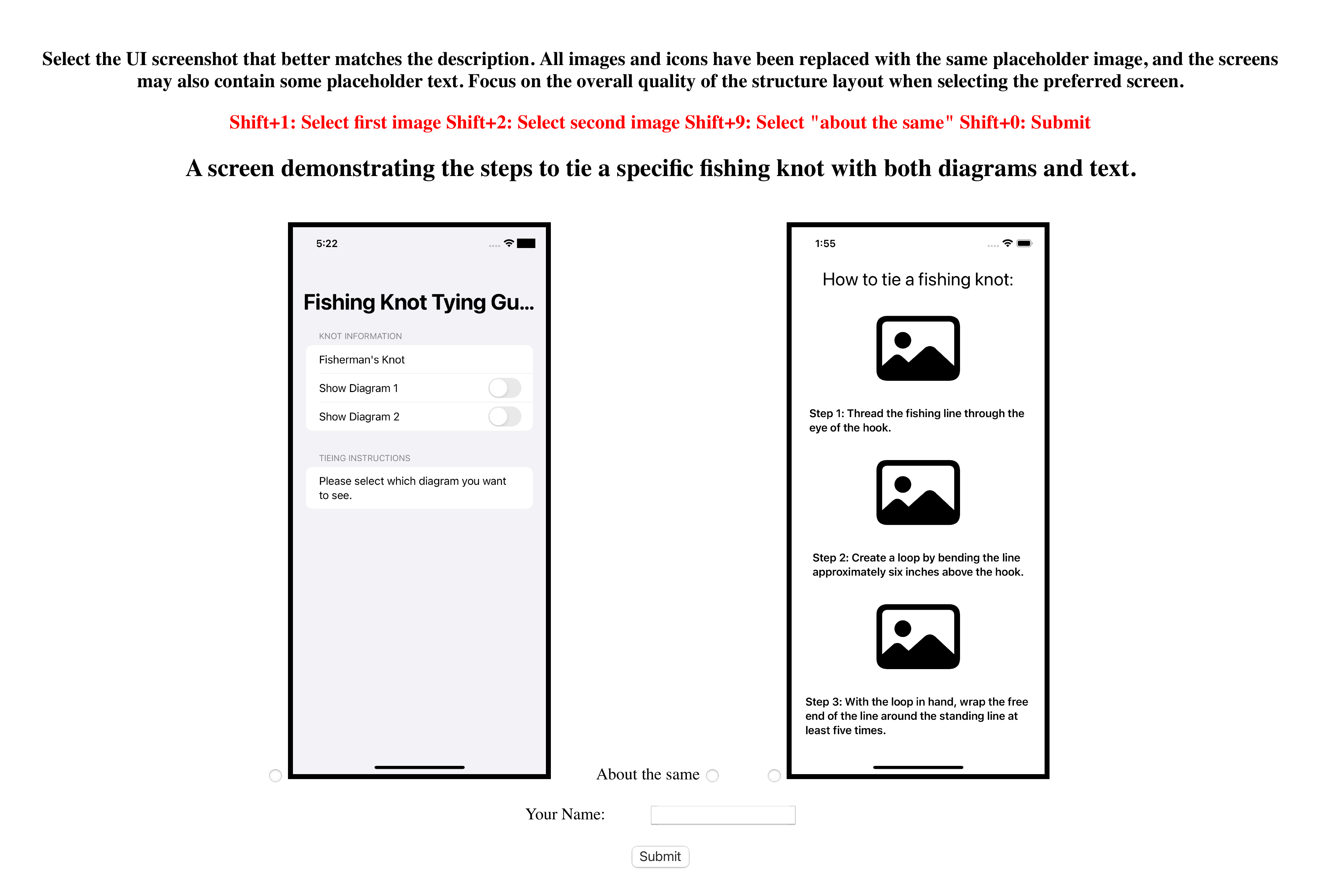}
  \caption{Screenshot of the interface presented to human evaluators to collect preferences rankings.}
  \label{fig:evalinterface}
\end{figure*}

We built a simple web application (Figure \ref{fig:evalinterface}) for collecting human preferences during our Elo evaluation. Our evaluation instructions asks the user to select the screenshot that better matches a description, which is randomly chosen from a list of 200 descriptions. We include the list of 200 input descriptions used for evaluation in supplementary material. The description is fed into two different models to generate two screenshots. The user can select one of the two screenshots or a third option indicating that they are ``about the same."

The full instructions are ``Select the UI screenshot that better matches the description. All images and icons have been replaced with the same placeholder image, and the screens may also contain some placeholder text. Focus on the overall quality of the structure and layout when selecting the preferred screen.''
\section{Hyperparameters}
\label{sec:appendix_training}
To aid in reproducability, we record hyperparameters used by various algorithms in Table \ref{tab:hyperparameters}. We did not conduct extensive hyperparameter sweeps, and set parameters based off default values and manual experimentation. It is possible that performance can be further improved through additional tuning.
Table \ref{tab:model_prompts} shows the various natural language prompts used in our training process and evaluation.
\begin{table}
\caption{Hyperparameters of models and algorithms used in our paper.}
\small
\begin{tabular}{@{}lll@{}}
\toprule
\textbf{Algorithm} & \textbf{Hyperparam.}      & \textbf{Value}                     \\ \midrule
UICoder         & Batch Size          & 1                         \\
                & Epochs              & 5                         \\
                & Learning Rate       & 1e-4                      \\
                & Top K               & 70                        \\
                & Top P               & 0.85                      \\
                & Temperature         & 0.2                       \\
                & Lora Rank           & 16                        \\
                & Lora Alpha          & 32                        \\
                & Lora Dropout        & 0.05                      \\
                & Lora Modules        & c\_proj, c\_attn, \\ & & q\_attn \\
                & Learning Rate (DPO) & 1e-5                      \\
                & Beta (DPO)          & 0.1                       \\
CLIP Filter     & Min Text Sim.       & 0.35                      \\
                & Min Visual Sim.     & 0.75                      \\
                & Percentile Thresh.  & 0.5\%                     \\
Dedup. Filter   & DBSCAN Eps.         & 0.25                      \\ \bottomrule
\end{tabular}
\label{tab:hyperparameters}
\end{table}
\begin{table*}[!htb]
\caption{Natural language prompts used for various models. When generating code with baseline models, our prompt was sometimes embedded inside of the model's own instruction-following prompt.}
\small
\begin{tabular}{@{}ll@{}}
\toprule
\textbf{Model}              & \textbf{Prompt}                                                                                                                                                                   \\ \midrule
BLIP-2             & screenshot of a mobile app showing <model completes text>                                                                                                                                     \\
Falcon-7B Instruct & rewrite the following description of a user interface for clarity "\{description\}". do not add any \\ &  additional details.                                                  \\
CLIP               & mobile user interface. well-designed. design awards winner. detailed app. featured screenshot. \\ &  \{description\}.                                                          \\
Code Generation    & Generate all required code that uses image assets and realistic placeholder data for a SwiftUI view \\ & named ContentView with the following description: "\{description\}." \\ \bottomrule
\end{tabular}
\label{tab:model_prompts}
\end{table*}

\end{document}